# PERIOCULAR BIOMETRICS: DATABASES, ALGORITHMS AND DIRECTIONS

*Fernando Alonso-Fernandez, Josef Bigun*

Halmstad University. Box 823. SE 301-18 Halmstad, Sweden.
{feralo, josef.bigun}@hh.se, http://islab.hh.se

## ABSTRACT

Periocular biometrics has been established as an independent modality due to concerns on the performance of iris or face systems in uncontrolled conditions. Periocular refers to the facial region in the eye vicinity, including eyelids, lashes and eyebrows. It is available over a wide range of acquisition distances, representing a trade-off between the whole face (which can be occluded at close distances) and the iris texture (which do not have enough resolution at long distances). Since the periocular region appears in face or iris images, it can be used also in conjunction with these modalities. Features extracted from the periocular region have been also used successfully for gender classification and ethnicity classification, and to study the impact of gender transformation or plastic surgery in the recognition performance. This paper presents a review of the state of the art in periocular biometric research, providing an insight of the most relevant issues and giving a thorough coverage of the existing literature. Future research trends are also briefly discussed.

## 1. INTRODUCTION

Periocular biometrics has been shown as one of the most discriminative regions of the face, gaining attention as an independent method for recognition or a complement to the face and iris modalities under non-ideal conditions [1, 2]. The typical elements of the periocular region are labeled in Figure 1 (top, left). This region can be acquired largely relaxing the acquisition conditions, in contraposition to the more carefully controlled conditions usually needed in face or iris modalities. Another advantage is that the problem of iris segmentation is avoided, which can be an issue in difficult images [3].

This paper provides a review of periocular research works, giving a framework which covers different aspects, including databases (Section 2) and algorithms for detection of the periocular region (Section 3). We also provide a taxonomy of features for periocular recognition (Section 4), which can be divided between those performing a *global* analysis of the image (extracting properties of an entire ROI) and those performing a *local* analysis (extracting properties of the neighborhood of a set of sparse selected key points). Besides personal recognition, other tasks have been also proposed using features extracted from the periocular region. In this direction, Section 5 deals with issues like soft-biometrics (gender/ethnicity classification), and impact of gender transformation and plastic surgery on the recognition accuracy.

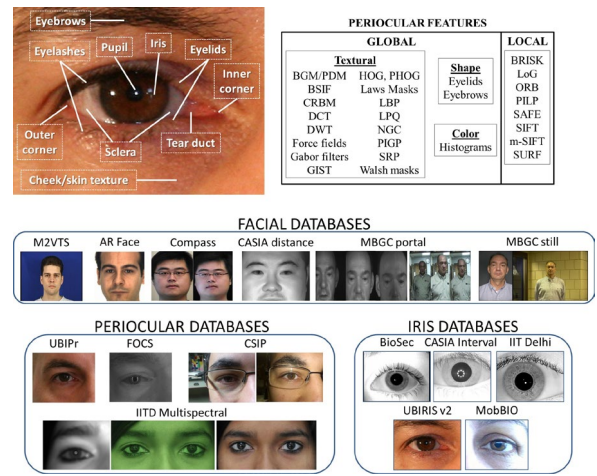

**Fig. 1**. Top (left): Elements of the periocular region. Top (right): Taxonomy of periocular features. Bottom: Samples of databases used in periocular research.

## 2. DATABASES FOR PERIOCULAR RESEARCH

Table 1 summarizes the databases used in periocular research, with some sample images in Figure 1, bottom. Acquisition setups include still images and videos with a variety of sensors in VW and NIR range such as: digital cameras, webcams, videocameras, smartphones, or close-up iris sensors. Although many databases have distance variability (e.g. FRGC, Compass, UBIRIS v2, UBIPr), acquisition is mostly done with the subject standing at several stand-off distances. A few number of databases contain video data of subjects walking through an acquisition portal (MBGC, FOCS), or

| Name | Subjects | Sessions | Data | Size | Illumination | Cross-spectr | Distance | Expression | Lightning | Occlusion | Pose | EER | Rank-1 |
|---|---|---|---|---|---|---|---|---|---|---|---|---|---|
| **FACIAL DATABASES** | | | | | | | | | | | | | |
| M2VTS [4] | 37 | 5 | 185 | 286×350 | V | N | N | Y | N | Y | Y | 0.3% | n/a |
| AR [5] | 126 | 2 | 4000 | 768×576 | V | N | N | Y | Y | Y | N | n/a | 76% |
| GTDB [6] | 50 | 2-3 | 750 | 640×480 | V | N | N | Y | Y | N | Y | 0.25% | 89.2% |
| Caltech [7] | 27 | n/a | 450 | 896×592 | V | N | N | Y | Y | N | N | 0.12% | n/a |
| FERET [8] | 1199 | 15 | 14126 | 512×768 | V | N | N | Y | Y | N | Y | 0.22% | 96.8% |
| CMU-H [9] | 54 | 1-5 | 764 | 640×480 | m | Y | N | N | Y | N | N | n/a | 97.2% |
| FRGC [10] | 741 | 1 | 36818 | 1200×1400 | V | N | Y | Y | Y | N | N | 0.09% | 98.3% |
| MORPH [11] | 515 | 2-5 | 1690 | 400×500 | V | N | N | Y | N | N | N | n/a | 33.2% |
| PUT [12] | 100 | n/a | 9971 | 2048×1536 | V | N | N | N | N | N | Y | 0.09% | 89.7% |
| MBGC v2 still [13] | 437 | n/a | 3482 | variable | V | N | N | Y | Y | N | N | 0.20% | 85% |
| MBGC v2 portal | 114 | n/a | 628 | 2048×2048 | N | Y | Y | N | Y | Y | N | 0.21% | 99.8% |
| | 91 | n/a | 571 | 1440×1080 | V | | | | | | | n/a | 98.5% |
| Plastic Surgery [14] | 900 | 2 | 1800 | 200×200 | V | N | N | N | N | N | N | n/a | 63.9% |
| ND-twins [15] | 435 | n/a | 24050 | 600×400 | V | N | N | Y | N | Y | Y | n/a | 98.3% |
| Compass [16] | 40 | 2 | 3200 | 128×128 | V | N | Y | Y | N | Y | Y | ~10% | n/a |
| FG-NET [17] | 82 | 12 | 1002 | 400×500 | V | N | Y | Y | Y | N | Y | 0.6% | 100% |
| CASIA Distance [18] | 142 | 1 | 2567 | 2352×1728 | N | N | N | N | N | N | N | n/a | 67% |
| FaceExpressUBI [19] | 184 | 2 | 90160 | 2056×2452 | V | N | N | Y | Y | N | N | 16% | n/a |
| **IRIS DATABASES** | | | | | | | | | | | | | |
| BioSec [20] | 200 | 2 | 3200 | 480×640 | N | N | N | N | N | N | N | 10.56% | 66% |
| CASIA Interval [18] | 249 | 2 | 2655 | 280×320 | N | N | N | N | N | N | N | 8.45% | n/a |
| UBIRIS v2 [21] | 261 | 2 | 11102 | 300×400 | V | N | Y | N | Y | Y | Y | 9.5% | 87.62% |
| IIT Delhi v1.0 [22] | 224 | 1 | 2240 | 240×320 | N | N | N | N | N | N | N | 1.88% | n/a |
| MobBIO [23] | 100 | 1 | 800 | 200×240 | V | N | N | N | Y | N | Y | 9.87% | 75% |
| **PERIOCULAR DATABASES** | | | | | | | | | | | | | |
| UBIPr [24] | 261 | 1-2 | 10950 | var. | V | N | Y | N | Y | Y | Y | 6.4% | 99.75% |
| FOCS [3] | 136 | var. | 9581 | 750×600 | N | N | Y | N | Y | Y | Y | 18.8% | 97.75% |
| IMP [25] | 62 | n/a | 620 | 640×480 | V | Y | Y | N | Y | N | N | 3.5% | n/a |
| | | | 310 | 600×300 | V | | | | | | | | |
| | | | 310 | 540×260 | n | | | | | | | | |
| CSIP [26] | 50 | n/a | 2004 | var. | V | Y | Y | N | Y | Y | Y | 15.5% | n/a |

'Illumination': V=VW, N=NIR, n=night, m=multispectral.
All databases have images, except M2VTS, CMU-H and MBGC v2 portal, which have videos.
'Best accuracy' is the best performance reported in the literature (Table 3).

**Table 1**. Public databases used in periocular research.

in hallways or atria (MBGC). There are databases captured to study particular problems too, such as aging (MORPH, FG-NET), plastic surgery, gender transformation, expression changes (FaceExpressUBI), face occlusion (AR, Compass), or cross-spectral matching (CMU-H, IMP).

Face and iris databases have been predominantly used for periocular research, since this region also appears in such databases. Iris images nevertheless have the limitation that eyebrows or other parts may not appear. Recently, some databases where the periocular region is specifically imaged and captured have been made public, such as FOCS (NIR portal), UBIPr (digital camera) or CSIP (smartphones). The 'best accuracy' shown in Table 1 should be taken as an approximate indication only, since different works may employ different subsets of the database or a different protocol. A general tendency, however, is that facial databases exhibit a better accuracy. These are the most used databases, so each new work builds on top of previous research, resulting in additional improvements. The use of databases acquired with personal devices such as smartphones or tablets is limited, with recognition accuracy still behind [26]. The same can be said about surveillance cameras [16]. Nevertheless, the accuracy with newer periocular databases are only some steps behind, demonstrating the capabilities of the periocular modality in difficult scenarios, where new research leaps are expected to bring accuracy to even better levels. New sensors are being proposed, such as Light Field Cameras, which capture multiple images at different focus in a single capture [27, 28], guaranteeing to have a good focused image.

## 3. PERIOCULAR DETECTION/SEGMENTATION

Initial periocular studies were focused on feature extraction only (with the periocular region manually extracted), but automatic detection and segmentation has increasingly become a research target in itself. Some works have applied a full face detector first such as the Viola-Jones (VJ) face detector [29], e.g. [30, 16], but successful extraction of the periocular region in this way relies on an accurate detection of the whole face. Using iris segmentation techniques for eye detection may not be reliable under challenging conditions either [3].

| Approach | Features | Database | Accuracy |
|---|---|---|---|
| [31] | Gabor filters | M2VTS | 99.3% |
| | | XM2VTS | 99% |
| [16] | ASM | Compass | n/a |
| [32] | VJ face parts | CASIA v4 distance | 96.4% |
| | | Yale-B (VW) | 99.2% |
| [33] | HSV + convex hull | UBIRIS v1 | n/a |
| [3] | Correlation filters | FOCS | 95% |
| [34] | LE-ASM + graph-cut | MBGC (VW still) | 99.4% |
| [35] | Correlation filters | HRT (VW images) | n/a |
| [36] | HSV | UBIRIS v1 | n/a |
| [37] | HSV+YCbCr | UBIRIS v2 / FRGC | n/a |
| [38] | Texture/shape | UBIRIS v2 | 97.5% |
| [39] | Symmetry filters | 4 NIR iris databases | 96% |
| | | 2 VW iris databases | 79% |
| [40] | VJ eyes + Hough | MBGC portal, UBIPr | n/a |
| | VJ eyes + morphology | CMU-H | n/a |

**Table 2**. Eye/periocular detection and segmentation works. Acronyms are defined in the text or referenced papers.

Table 2 summarizes existing research aimed at locating the eye position directly. Eye detection can be a decisive preprocessing to ensure successful iris segmentation in difficult images, as in [3] using correlation filters over the difficult FOCS database. Correlation filters were also used in [35], but after applying the VJ face detector. [32] and [40] used the VJ detector of face sub-parts. [40] also experimented with the CMU hyperspectral database, which has images captured simultaneously at multiple wavelengths. Since the eye is centered in all bands, accuracy can be boosted by collective detection over all bands. [31] made use of Gabor features for eye detection and face tracking purposes by performing saccades across the image, whereas [41, 39] used symmetry filters tuned to detect circular symmetries. The latter has the advantage of not needing training, and detection is possible with a few 1D convolutions due to separability of the detection filters, built from derivatives of Gaussians. [34] proposed a Local Eyebrow Active Shape Model (LE-ASM) to detect the eyebrow region directly from a given face image, with eyebrow pixels segmented afterwards using graph-cut segmentation. ASMs were also used by [16] to automatically extract the periocular region, albeit after the application of a VJ face detector. Recently, [38] proposed a method to label 7 components of the periocular region by using 7 classifiers at the pixel level, with each classifier specialized in one component. Pixel features included texture and shape descriptors. Some works have proposed the extraction of features from the sclera region only. For this purpose, [36], [37] and [33] have used the HSV/YCbCr color spaces for sclera segmentation. In these works, however, sclera detection is guided by a prior detection of the iris boundaries.

## 4. FEATURES FOR PERIOCULAR RECOGNITION

Features employed for periocular recognition can be classified into global and local approaches (Figure 1, top, right). *Global approaches* extract properties of an entire region of interest (ROI), such as texture, shape or color features. In *local approaches*, on the other hand, a sparse set of characteristic points (called key points) is detected first, with features describing the neighborhood around key points only. Some pre-processing steps have been also used to cope with the difficulties found in unconstrained scenarios, such as pose correction by Active Appearance Models (AAM) [42], illumination normalization [43, 44], correction of deformations due to expression change by Elastic Graph Matching (EGM) [45], or color device-specific calibration [26]. Table 3 gives an overview of existing works for automatic periocular recognition. The most widely used approaches include Local Binary Patterns (LBP) and, to a lesser extent, Histogram of Oriented Gradients (HOG) Transform (SIFT) key points. Over the course of the years, many other descriptors have been proposed. The following is a brief review of these works, highlighting the most important results or contributions. For the sake of pages limitation, we omit the description of the feature extraction techniques, or references to the original works where they are presented. There are also periocular studies evaluating the ability of (untrained) human observers to compare pairs of periocular images both with VW and NIR illumination [46, 47].

Periocular recognition started to gain popularity after the studies by [48, 30]. Some pioneering works can be traced back to 2002 [31], although authors here did not call the local eye area 'periocular'. The good performance reported by [30] set the framework for the use of the periocular modality for recognition. Many works have followed this approach as inspiration, either improving it or introducing new perspectives, with LBPs and their variations being particularly extensive in the literature [49, 50, 51, 52, 53, 54]. Some works have also employed other features in addition to these [52, 54]. The use of subspace representation methods after LBP extraction is also becoming a popular way either to improve performance or reducing the feature set [40, 43, 55, 56, 57].

Inspired by [48], [58] extended the experiments with additional global and local features, to a significant larger set of the FRGC database with less ideal images (thus the lower accuracy w.r.t. previous studies). They later addressed the problem of aging degradation on periocular recognition [42], reported to be an issue even at small time lapses [30]. To obtain age invariant features, they first performed preprocessing schemes, such as pose correction, illumination and periocular region normalization. LBP has been also used in other works analyzing for example the impact of plastic surgery [59] or gender transformation [35] on periocular recognition (see Section 5).

The framework set by [31] with Gabor filters served as inspiration for experiments with several databases in NIR and VW [60, 41, 39]. Authors later proposed a matcher based on Symmetry Assessment by Feature Expansion (SAFE) descriptors [61, 39], which describes neighborhoods around key-points by estimating the presence of various symmetric curve families. Gabor filters were also used by [62] in their work presenting Local Phase Quantization (LPQ) as periocular descriptor. [63] also employed Gabor features over four different VW databases, with features reduced by Direct Linear Discriminant Analysis (DLDA) and further classified by a Parzen Probabilistic Neural Network (PPNN).

[64] evaluated Circular Local Binary Patterns (CLBP) and a global descriptor (GIST) consisting of five perceptual dimensions related with scene description (image naturalness, openness, roughness, expansion and ruggedness). They used the UBIRIS v2 database of uncontrolled VW iris images which includes a number of perturbations intentionally introduced (see Section 2). A number of subsequent works have also made use of UBIRIS v2 [65, 66, 67, 37, 38, 68]. [67] proposed Phase Intensive Global Pattern (PIGP) features, and they later proposed the Phase Intensive Local Pattern (PILP) feature extractor [68], reporting the best Rank-1 periocular performance to date with UBIRIS v2. [37] extracted features from the eyelids region only, and [38] proposed a method to label seven components of the periocular region (see Section 3) with the purpose of demonstrating that regions such as hair or glasses should be avoided since they are unreliable for recognition. Finally, [36] used the first version of UBIRIS in their study presenting directional projections or Structured Random Projections (SRP) as periocular features.

Other shape features have been also proposed, like eyebrow shape features [69, 34], with surprisingly accurate results when the eyebrow is used as a stand-alone trait. Indeed, eyebrows have been used by forensic analysts for years to aid in facial recognition [34], suggested to be the most salient and stable features in a face [70]. [69] also used the extracted eyebrow features for gender classification (see Section 5). [34] proposed a eyebrow segmentation technique too (Section 3).

[24] presented the first periocular database specifically acquired for periocular research (UBIPr). They also proposed to compute the ROI w.r.t. the midpoint of the eye corners (instead of the pupil center), which is less sensitive to gaze variations, leading to a significant improvement (EER from ∼30% to ∼20%). Posterior studies have managed to improve

| | Features evaluated | Test Database | Best accuracy EER | Best accuracy Rank-1 | |
|---|---|---|---|---|---|
| [31] | Gabor filters | M2VTS | 0.3% | n/a | |
| [48] | HOG, LBP, SIFT | FRGC | 6.96% | 79.49% | |
| [49] | GEFE+LBP | FRGC | n/a | 86.85% | |
| | | FERET | n/a | 80.80% | |
| [50] | CLBP, GIST | UBIRIS v2 | n/a | 63.34% | |
| [46] | Human observers | Prop. (NIR) | n/a | 92% | |
| [51] | LBP, WLBP, SIFT, DCT, DWT | FRGC | n/a | 53.2% | * |
| [42] | SURF, Walsh/Law masks | FG-NET | 0.6% | 100% | |
| [42] | Gabor, Force Fields, LoG | | | | |
| [52] | LBP | FRGC | 0.10% | 84.39% | |
| | | FERET | 0.22% | 72.22% | |
| [53] | LBP | MBGC (NIR portal) | 21% | 92.5% | |
| [54] | RG color histogram, LBP | FRGC | n/a | 96.8% | |
| | | MBGC (NIR portal) | n/a | 87% | |
| [55] | BGM | FOCS | 23.81% | 94.2% | |
| [56] | eyebrow shape | MBGC (NIR portal) | n/a | 91% | |
| | | FRGC | n/a | 78% | |
| [41] | Gabor filters | BioSec | 10.56% | 66% | |
| [39] | | CASIA | 14.53% | n/a | |
| | | IITD | 2.5% | n/a | |
| | | MobBIO | 12.32% | 75% | |
| | | UBIRIS v2 | 24.4% | n/a | |
| [47] | Human observers | Prop. (VW) | n/a | 88.4% | |
| | | Prop. (NIR) | n/a | 78.8% | |
| [57] | SIFT, LBP | Plastic Surgery | n/a | 63.9% | * |
| [58] | LBP | UBIRIS v2 | 12.94% | 81.03% | |
| [16] | WLBP | Compass | ∼10% | n/a | * |
| [59] | LBP, PCA/LDA variants | FERET | ∼15% | n/a | |
| [24] | HOG, LBP, SIFT | UBIPr | ∼20% | n/a | |
| [60] | HOG, m-SIFT, PDM | FOCS | 18.8% | n/a | |
| | | FRGC | 1.59% | n/a | |
| [61] | LBP, SIFT | UBIRIS v2 | 31.87% | 56.4 | |
| [62] | SIFT, LBP, HOG, LMF | CASIA v4 Distance | n/a | ∼67% | |
| [63] | LBP, 3PLBP, H3PLBP | Morph | n/a | 33.2% | * |
| | | FRGC | n/a | 97.51% | * |
| | | Georgia Tech | | 92.4% | |
| | | ND Twins | | 98.03% | |
| [27] | LBP+SRC | Prop. light-field cam | 12.04% | n/a | |
| | | Prop. digital cam | 16.21% | n/a | |
| [64] | PDM, m-SIFT | FOCS | 18.85% | 97.75% | |
| | | UBIPr | 6.43% | 99.75% | |

| | Features evaluated | Test Database | Best accuracy EER | Best accuracy Rank-1 | |
|---|---|---|---|---|---|
| [65] | raw pixels, LBP, PCA, LBP+PCA | MGBC (NIR portal) | n/a | 97.7% | * |
| [66] | PIGP, CLBP, WLBP | UBIRIS v2 | n/a | 82.86% | |
| [67] | LPQ, LBP, Gabor filters | Caltech | 0.12% | n/a | |
| | | PUT | 0.09% | | |
| | | GTDB | 0.28% | | |
| | | MBGC (VW still) | 0.22% | | |
| [68] | LBP, NGC, JDSR | Propr. iris (NIR), face (VW) | 6% | n/a | |
| [69] | Gabor-PPNN, DWT, LBP, HOG | MBGC (VW still) | 6.4% | 75.8% | * |
| | | GTDB | 5.9% | 89.2% | * |
| | | IITK | 15.5% | 67.6% | |
| | | PUT | 4.8% | 89.7% | |
| [70] | SIFT, SURF, BRISK, ORB, LBP | FERET | n/a | 96.8% | |
| [34] | Eyebrow shape | MBGC (VW still) | n/a | 85% | * |
| | | AR | n/a | 76% | * |
| [35] | TPLBP, LBP, HOG | HRT | 35.21% | 57.79% | * |
| [71] | Symmetry patterns (SAFE) | BioSec | 10.75% | n/a | |
| [72] | | CASIA | 8.45% | | |
| | | IITD | 1.88% | | |
| | | MobBIO | 9.87% | | |
| | | UBIRIS v2 | 24.56% | | |
| [44] | PCA to: CRBM, DSIFT, LBP, HOG | UBIPr | 6.4% | 50.1% | |
| [36] | Directional projections (SRP) | UBIRIS v1 | 6.52% | n/a | |
| [37] | LBP, eyelids shape | FRGC | <25% | n/a | |
| | | UBIRIS v2 | <24% | n/a | |
| [45] | GC-EGM to: LBP+HOG+SIFT | FaceExpressUBI | 16% | n/a | |
| [38] | LBP, HOG, SIFT | UBIRIS v2 | 9.5% | n/a | |
| [28] | BSIF | Propr. light-field cam | 3.39% | n/a | |
| | | Propr. digital cam | 3.96% | n/a | |
| [26] | LBP, HOG, SIFT, CLBP, GIST | CSIP | 15.5% | n/a | |
| [25] | LBP, HOG, PHOG | IMP (VW) | ∼8% | n/a | * |
| | FPLBP, PHOG+NN | IMP (night) | ∼7% | n/a | * |
| | | IMP (NIR) | ∼3.5% | n/a | * |
| [73] | PILP, SIFT, SURF | Bath | n/a | 100% | |
| | | CASIA Lamp v3 | | 100% | |
| | | UBIRIS v2 | | 87.62% | |
| | | FERET | | 85.8% | |
| [40] | raw pixels, LBP, PCA, LBP+PCA | MGBC (NIR portal) | n/a | 99.8% | |
| | | MGBC (VW portal) | | 98.5% | |
| | | CMU | | 97.2% | |
| | | UBIPr | | 99.5% | |

Best accuracy refers to single eye matching, except rows marked with asterisk (*).

**Table 3**. Existing periocular recognition works in chronological order (from top to bottom and left to right). Acronyms are fully defined in the text or referenced papers.

performance over the UBIPr database using a variety of features [71, 44]. The UBIPr database is also used by [40] in their extensive study evaluating data in VW (UBIPr, MBGC), NIR (MBGC) and multi-spectral (CMU-H) range, with the reported Rank-1 results being the best published performance to date for the four databases employed. A new database of challenging periocular images (CSIP) was presented recently by [26], the first one made public captured with smartphones. Another database captured specifically for cross-spectral periocular research (IMP) was also recently presented by [25], containing data in VW, NIR and night modalities. Cross-spectral recognition was also addressed by [72] using a proprietary database. Finally, [27] and [28] presented a database in VW range acquired with a new type of camera, a Light Field Camera (LFC), which provides multiple images at different focus in a single capture. LFC overcomes one important disadvantage of sensors in VW range, which is guaranteeing a good focused image. Unfortunately, the database has not been made available. Individuals were also acquired with a conventional digital camera, with a superior performance observed with the LFC camera. New features were also presented in the two studies.

## 5. SOFT-BIOMETRICS, GENDER TRANSFORMATION AND PLASTIC SURGERY.

Besides personal recognition, a number of other tasks have been also proposed using features from the periocular region, as shown in Table 4. *Soft-biometrics* refer to the classification of an individual in broad categories such as gender, ethnicity, age, height, weight, hair color, etc. While these cannot be used to uniquely identify a subject, it can reduce the search space or provide additional information to boost the recognition performance. Due to the popularity of facial recognition, face images have been frequently used to obtain both gender and ethnicity information, with high accuracy (>96%, for a summary see [73]). Recently, it has been also suggested that periocular features can be potentially used for soft-biometrics [74, 73, 75, 76], with accuracies comparable to these obtained by using the entire face. [75] also showed that fusion of the soft-biometrics information with texture features from the periocular region can improve the recognition performance [75]. An interesting study by [69] with features from the eyebrow region only showed very good results in gender classification.

Other studies are related with the effect on the recognition performance of plastic surgery or gender transformation. [35] studied the impact of gender transformation via Hormone Replacement Theory (HRT), which causes changes in the physical appearance of the face and body gradually over the course of the treatment. A database of >1.2 million face images from YouTube videos was built, observing that accuracy of the periocular region greatly outperformed other face components (nose, mouth) and the full face. Also, face matchers began to fail after only a few months of HRT treatment. [59] studied the matching of face images before and after undergoing plastic surgery, outperforming previous studies where only full-face matchers were used. Both [35] and [59] used Commercial Off The Shelf (COTS) full face recognition systems too (VeriLook, PittPatt and FaceVACS).

| | Aim | Features | Database | Best accuracy |
|---|---|---|---|---|
| [76] | GC | raw pixels, LBP + LDA-NN/PCA-NN/SVM | Prop. (936 VW images) | GC: 85% |
| [69] | GC | eyebrows shape + MD/LDA/SVM | FRGC MBGC NIR portal | GC: 97% GC: 96% |
| [59] | PS | Periocular: SIFT, LBP Face: VeriLook (VL) Face: PittPatt (PP) | Plastic Surgery | Rank-1 periocular: 63.9%% Rank-1 face: 85.3% Rank-1 periocular+face: **87.4%** |
| [74] | GC | ICA + NN | FERET | GC: 90% |
| [73] | GC EC | LBP/HOG/DCT/LCH + ANN/SVM | FRGC MBGC NIR portal | GC: 97.3%, EC: 94% GC: 90%, EC: 89% |
| [35] | GT | LBP, TPLBP, HOG applied to face parts and to the full face Face: PittPatt (PP) Face: FaceVACS (FV) | HRT (>1.2 million VW images) | Perioc: EER=**35.21%**, Rank-1=**57.79%** Nose: EER=41.82%, Rank-1=44.57% Mouth: EER=43.25%, Rank-1=39.24% Face: EER=38.6%, Rank-1=46.69% PP: EER=n/a, Rank-1=36.99% FV: EER=n/a, Rank-1=29.37% |

GC: Gender Classification. EC: Ethnicity Classification. PS: Plastic Surgery. GT: Gender Transformation.
GC, EC works report classification rates while PS, GT report recognition rates before/after transformation.

**Table 4**. Existing works on soft-biometrics, gender transformation and plastic surgery analysis using periocular features. Acronyms are fully defined in the text or referenced papers.

## 6. CONCLUSIONS

This paper reviews existing literature in periocular biometrics research. Our aim is to provide an insight of the most relevant issues and challenges. The fast-growing uptake of face technologies in social networks and smartphones, as well as the widespread use of surveillance cameras, arguably increases the interest of this modality. The periocular region has shown to be more tolerant to variability in expression, occlusion, and it has more capability of matching partial faces [43]. It also finds applicability in other areas of forensics analysis (crime scene images where perpetrators intentionally mask part of their faces). In such situation, identifying a suspect where only the periocular region is visible is one of the toughest real-world challenges in biometrics. Even in this difficult case, the periocular region can aid in the reconstruction of the whole face [77].

Periocular biometrics has rapidly evolved to competing with face or iris recognition. Under difficult conditions, such as acquisition portals, [50, 78, 79], distant acquisition, [52], smartphones, [26], webcams or digital cameras, [39, 80], the periocular modality is shown to be clearly superior to the iris modality, mostly due to the small size of the iris or the use of visible illumination. The latter is predominant in relaxed or uncooperative setups due to the impossibility of using NIR illumination. Iris texture is more suited to the NIR spectrum, since this type of lightning reveals the details of the iris texture [81], while the skin reflects most of the light, appearing over-illuminated. On the other hand, the skin texture is clearly visible in VW range, but only irises with moderate levels of pigmentation image reasonably well in this range [82]. However, despite the poor performance shown by the iris in the visible spectrum, fusion with periocular is shown to improve the performance in many cases [66, 80]. Similar trends are observed with face. Under difficult conditions, such as blur or downsampling, the periocular modality performs considerably better [83]. It is also the case of partial face occlusions, where performance of full-face matchers is severely degraded [30]. The periocular modality is also shown to aid or outperform face matchers in case of undergoing plastic surgery [59] or gender transformation [35].

Another issue that is receiving increasing attention is cross-modality [72], cross-spectral [84, 25], hyper-spectral [40] or cross-sensor [26] matching. The periocular modality also has the potential to allow ocular recognition at large stand-off distances [84], with applications in surveillance. Samples captured with different sensors are to be matched if, for example, people is allowed to use their own smartphone (cross-sensor), not to mention if the sensors work in different spectral range (cross-spectral). The same occurs with data captured with surveillance cameras. Exchange of biometric information between different law enforcement agencies worldwide also poses similar problems. These are examples of some scenarios where, if biometrics is extensively deployed, data acquired from heterogeneous sources will have to co-exist [85], These issues are of high interest in new scenarios arising from the widespread use of biometric technologies and the availability of multiple sensors and vendor solutions.